\documentclass[letterpaper]{article} 
\usepackage{aaai2026}  
\usepackage{times}  
\usepackage{helvet}  
\usepackage{courier}  
\usepackage[hyphens]{url}  
\usepackage{graphicx} 
\usepackage{natbib}  
\usepackage{caption} 
\usepackage{algorithm}
\usepackage{algorithmic}

\usepackage{pifont}
\usepackage{amsmath}
\usepackage{amssymb}
\usepackage{booktabs}
\usepackage{multirow}

\usepackage{newfloat}
\usepackage{listings}
\DeclareCaptionStyle{ruled}{labelfont=normalfont,labelsep=colon,strut=off} 
\floatstyle{ruled}
\newfloat{listing}{tb}{lst}{}
\floatname{listing}{Listing}
\title{Multivariate Gaussian Representation Learning for Medical Action Evaluation}
\author{
    Luming Yang\textsuperscript{\rm 1}\equalcontrib\footnotemark[2], Haoxian Liu\textsuperscript{\rm 2}\equalcontrib, Siqing Li\textsuperscript{\rm 3}, Alper Yilmaz\textsuperscript{\rm 1}\thanks{Correspondence to \{yilmaz.15, yang.7670\}@osu.edu}\\
}
\affiliations{
    \textsuperscript{\rm 1}The Ohio State University\\
    \textsuperscript{\rm 2}Hong Kong University of Science and Technology\\
    \textsuperscript{\rm 3}Southern University of Science and Technology\\
    yang.7670@osu.edu,
    hliueu@connect.ust.hk,
    lisq2022@mail.sustech.edu.cn,
    yilmaz.15@osu.edu
}

\begin{document}

\maketitle

\begin{abstract}
Fine-grained action evaluation in medical vision faces unique challenges due to the unavailability of comprehensive datasets, stringent precision requirements, and insufficient spatiotemporal dynamic modeling of very rapid actions. 
To support development and evaluation, we introduce \textsc{CPREval-6k}, a multi-view, multi-label medical action benchmark containing 6,372 expert-annotated videos with 22 clinical labels.
Using this dataset, we present \textsc{GaussMedAct}, a multivariate Gaussian encoding framework, to advance medical motion analysis through adaptive spatiotemporal representation learning. Multivariate Gaussian Representation projects the joint motions to a temporally scaled multi-dimensional space, and decomposes actions into adaptive 3D Gaussians that serve as tokens. These tokens preserve motion semantics through anisotropic covariance modeling while maintaining robustness to spatiotemporal noise. Hybrid Spatial Encoding, employing a Cartesian and Vector dual-stream strategy, effectively utilizes skeletal information in the form of joint and bone features.
The proposed method achieves 92.1\% Top-1 accuracy with real-time inference on the benchmark, outperforming the baseline by +5.9\% accuracy with only 10\% FLOPs. Cross-dataset experiments confirm the superiority of our method in robustness.
\end{abstract}

\begin{links}
    \link{Code}{https://github.com/HaoxianLiu/GaussMedAct}
\end{links}

\section{Introduction}

Cardiac arrest claims over 436,000 lives annually in the US alone, where high-quality cardiopulmonary resuscitation (CPR) can double survival rates.
Recent advances in medical vision systems have underscored the critical need for fine-grained and rapid motion understanding in time-sensitive clinical scenarios, particularly CPR \cite{patil2015cardiac, masterson2024real}. The quality of CPR directly determines survival outcomes in cardiac arrest emergencies \cite{schultz2015strategies}, where compression depth and frequency are strongly correlated with survival rate \cite{schultz2015strategies, daudre2023evaluating}. Using feedback techniques can effectively improve results \cite{yeung2009use}. However, the current manual CPR assessment suffers from fundamental limitations revealed in our experiments. The results showed that human evaluators achieve only 74.8\% accuracy in detecting critical errors such as incomplete chest recoil and frequency deviations (tested in 23 certified practitioners). Existing computer vision systems fail to capture centimeter motion deviations (\textit{e.g.} $5$ cm compression depth \cite{stiell2012role}) and millisecond-level frequency variations (\textit{e.g.} $115\pm5$ bpm requirements) \cite{travers2010part}. These challenges stem from persistent gaps in visual computing: modeling anatomical causality in spatiotemporal dynamics, preserving clinical interpretability, and achieving medical-grade temporal precision.

\begin{figure}
    \centering
    \includegraphics[width=1\linewidth]{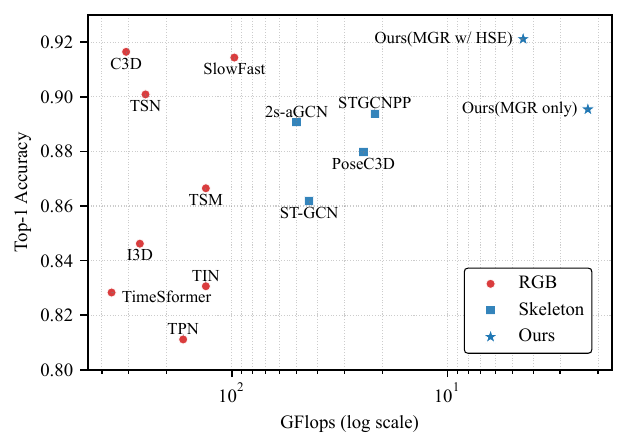}
    \caption{\textbf{Efficiency-Accuracy Trade-off Comparison.} Each point represents a model, with coordinates indicating computational complexity (GFLOPs) in log scale and top-1 accuracy. The proposed model is Pareto optimal.}
    \label{fig:baseline}
\end{figure}

\begin{figure*}
    \centering
    \includegraphics[width=\linewidth]{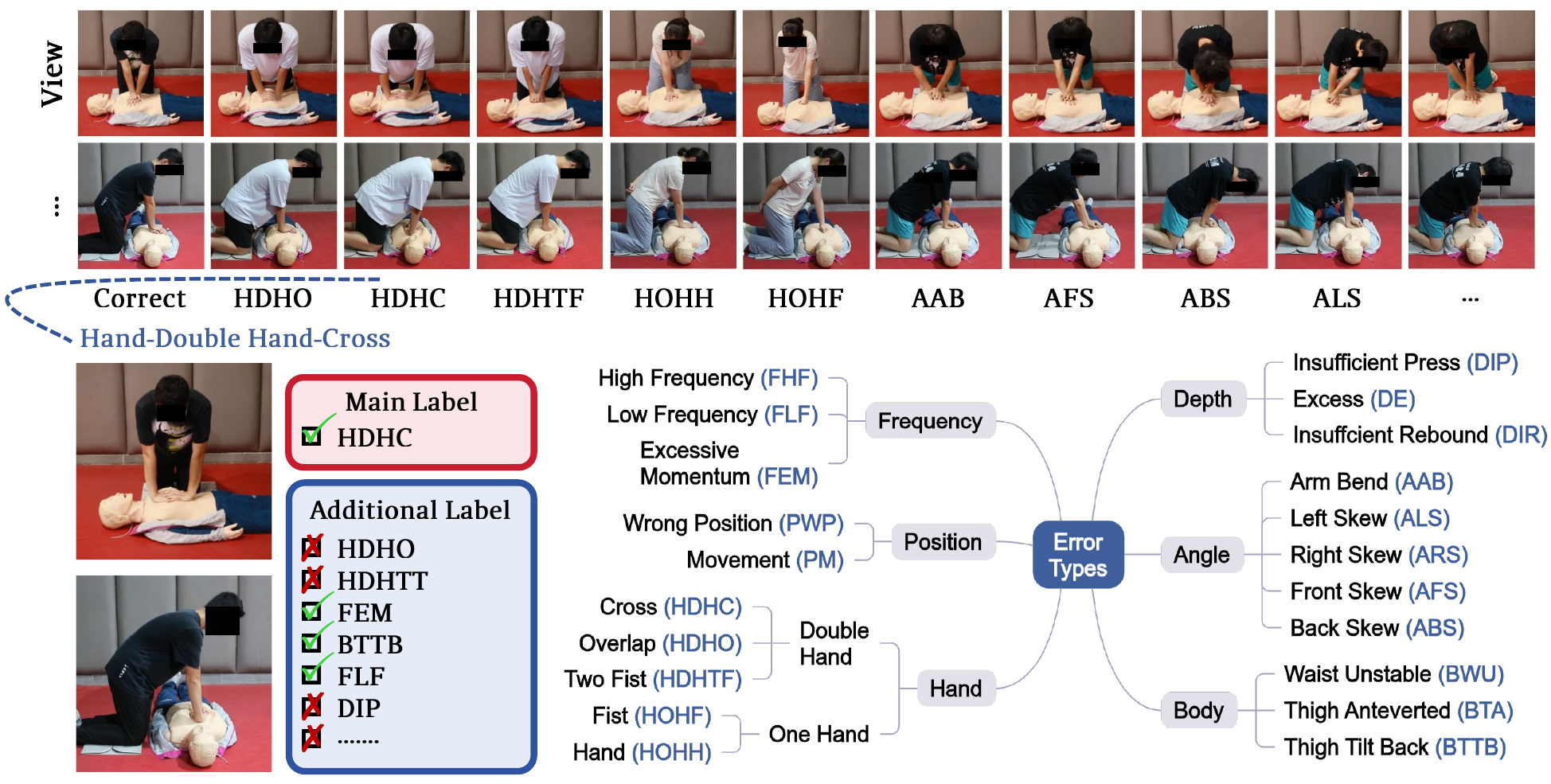}
    \caption{\textbf{Dataset Overview.} A multi-view CPR dataset with hierarchical error annotation, comprising 6 primary error classes and 21 fine-grained sub-classes. Each instance includes one primary error label and multiple secondary labels for compound error analysis.}  
    \label{fig:dataoverview}
\end{figure*}

Current action recognition approaches that focus on images (RGB-based) or graphs (skeleton-based) face limitations. Methods based on RGB (\textit{e.g.} TimeSformer \cite{bertasius2021space}) that rely on pre-trained backbones suffer from drawbacks such as fundamentally lacking anatomical modeling capabilities while incurring prohibitive computational latency. Skeleton-based methods (\textit{e.g.} ST-GCN \cite{yu2018spatio}) discard motion semantics through rigid temporal pooling operations and remain vulnerable to noise. A good solution requires a balance between accuracy, explainability, and latency \cite{gruber2012real}. These limitations directly affect the reliability of any method in the real world, while suboptimal CPR guidance techniques do not improve lifesaving effectiveness \cite{nassar2017improving}. 

Recent advances in Gaussian Splatting have demonstrated efficient and high-quality rendering in computer graphics, suggesting that sparse Gaussian distributions can effectively represent complex 3D point clouds \cite{kerbl20233d, chen2024survey}. However, its potential remains unexplored for spatiotemporal representations. This gap presents an opportunity given two key observations: First, Gaussian Mixture Models \cite{reynolds2009gaussian} naturally align with the probabilistic nature of motion point-cloud distributions. Second, conjecture on spatiotemporal interest points reveals that human motion dynamics can be decomposed into temporal components \cite{laptev2005space}. Building on these insights, we propose a Multivariate Gaussian Representation (MGR) for robust spatiotemporal skeleton learning, which models temporal evolution of keypoints as probability distributions to achieve compact and noise-resistant action representations.

Our design is further motivated by fundamental principles of motion perception. Psychological studies show that sparse 2D point-light displays can convey strong action impressions through basic kinematic patterns \cite{johansson1973visual}. Complementary motion semantics exist in both absolute joint positions and relative bone kinematics \cite{shi2019two}. To fully exploit this dual nature, we introduce a Hybrid Spatial Encoding (HSE) through a Cartesian-Vector dual-stream architecture reconciles absolute anatomical positioning with relative kinematic patterns - an approach particularly effective for modeling rapid human motion.

Our key contributions can be summarized as follows:

\begin{itemize}
    \item \textsc{CPREval-6k}: The largest multi-view clinical CPR dataset featuring synchronized RGB-skeleton streams and expert-validated multi-label annotations. Our benchmark contains 6,372 chest compression clips in 22 categories, each video hierarchically annotated with a primary critical error and secondary factors.
    \item \textsc{GaussMedAct}: An end-to-end framework that combines MGR and HSE. By generating precision action token tensors, \textsc{GaussMedAct} enables multiple downstream tasks including real-time classification and the generation of evaluation reports.
\end{itemize}

\begin{figure*}
    \centering
    \includegraphics[width=\linewidth]{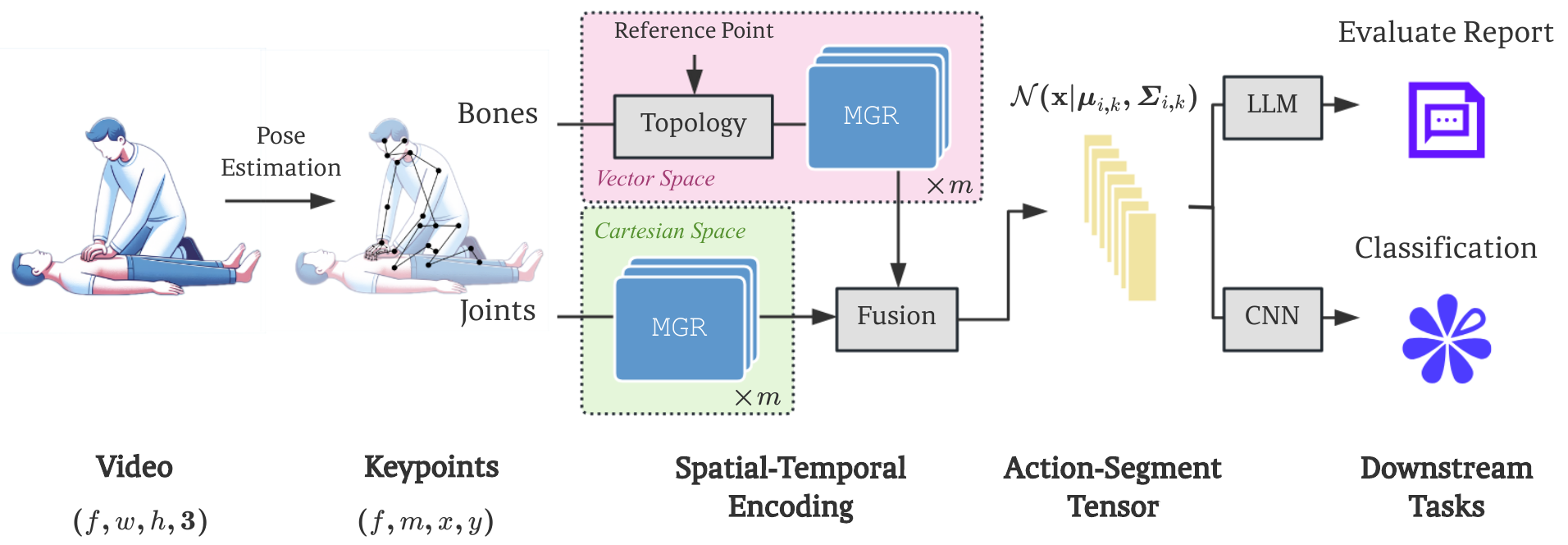}
    \caption{\textbf{Schematic of the \textsc{GaussMedAct} Pipeline.} Input data undergoes cartesian and vector based dual-stream encoding and pass through MGR to generate gaussians. Through feature fusion, action tokens are generated for downstream tasks.}
    \label{fig:pipeline}
\end{figure*}

\section{CPREval-6k Dataset}
\label{sec:cpreval}
The proposed dataset \textsc{CPREval-6k} comprises 6,372 manually annotated videos documenting manual chest compression procedures, captured from multiple viewpoints. Through consultations with emergency physicians and team expertise, we identified 21 different error categories in 6 medically critical aspects of chest compression techniques, including `hand position', `arm angle', `body posture', `compression depth', `operation frequency', and `body positioning', as shown in Figure \ref{fig:dataoverview}. Each CPR video is annotated with one primary error designation and multiple secondary error labels. See supplementary material for details of the dataset.

\noindent\textbf{Data Collection.}
We implemented a multiview camera recording system with synchronized initiation to ensure temporal alignment between perspectives. To validate dataset robustness, 6 volunteers who had not received professional CPR training and 6 American Heart Association (AHA) Heartsaver$^\circledR$ CPR-certified volunteers were recruited for standardized error simulation. The participants performed equal repetitions of each predefined error category, establishing a balanced distribution for the experimental validity. The entire dataset collection process is under the supervision of professionals to ensure that their actions meet the data collection standards.

\noindent \textbf{Hierarchical Annotation.} During data verification, we observed the involuntary co-occurrence of secondary errors during primary error simulations, necessitating our hierarchical annotation scheme of primary-secondary error labeling. A quality assurance protocol (QAP) involved 10 certified annotators who passed inter-rater reliability testing through rigorous Inter-Annotator Agreement evaluation. Annotation guidelines were strictly enforced under the supervision of a certified instructor who conducted final label verification, ensuring consistency and eliminating conspicuous annotation errors.

\noindent\textbf{Data Analysis.} 
The annotated dataset exhibits a significant class imbalance in the primary labels. Among the primary error labels, the \textit{Depth-Insufficient Press} type accounts for 9.4\%, making it the most frequent primary error type. This distribution arises from the Explicit Error Prioritization annotation protocol, where each sample is assigned only one dominant primary error along with multiple secondary errors. 

In the overall error category distribution, which includes both primary and secondary error labels, \textit{Depth-Insufficient Press} emerges as the most prevalent error type, accounting for 29.1\% of all annotated error instances. This is consistent with clinical studies~\cite{bobrow2013influence} that highlight challenges in monitoring chest compression quality.

The analysis of secondary labels under the primary label reveals frequent co-occurrence patterns between specific error types. Preliminary correlation analysis identifies statistically significant relationships. For example, \textit{Freq-Excessive Momentum} and \textit{Depth-Excess} exhibit a Pearson correlation coefficient of 0.52, while \textit{Wrong Position} and \textit{Position-Movement} show a correlation of 0.39, which aligns with biomechanical intuition.

To systematically uncover hierarchical error dependencies, we implement an enhanced Apriori algorithm ($min\_support=0.025$, $min\_confidence=0.25$) for the mining of association rules. The results are listed in the supplementary material. We have the following key findings:

\begin{itemize}
    \item \textbf{Strong Association:} Rule 5 (M:\textit{Position-Movement} $\rightarrow$ A:\textit{Position-Wrong Position}) demonstrates an exceptional association, achieving 77.6\% confidence and $17.4\times$ lift. This quantifies the linkage where compression point instability directly induces positional errors.
    \item \textbf{Kinematic Chain Coupling:} Upper-limb anomalies (M:\textit{Angle-Arm Bend}) correlate with both posterior thigh tilt (A:\textit{Body-Thigh Tilt Back}, 38.8\% confidence) and insufficient compression depth (A:\textit{Depth-Insufficient Press}, 38.4\% confidence), revealing whole-body kinetic chain interactions during CPR execution.
    \item \textbf{Hierarchical Error Propagation:} M$\rightarrow$A rules exhibit higher confidence (38.8--77.6\%) compared to A$\rightarrow$A co-occurrence rules (26.1--32.7\%), indicating that primary errors exert strong causative drives on secondary errors.
\end{itemize}

Statistical data indicate that manual chest compressions constitute a coordinated technical action of the entire body. A single error in execution can propagate additional errors, ultimately resulting in suboptimal CPR outcomes. This highlights the critical importance of mastering the correct posture of CPR to ensure effective performance.

\section{GaussMedAct}
We propose \textsc{GaussMedAct}, a dual-stream spatiotemporal encoding framework for medical action recognition (see Figure \ref{fig:pipeline}). The overall pipeline is summarized as follows.
\begin{itemize}
    \item \textbf{Spatial dimension.} Human key points are extracted using pose estimation, and their characteristics are decoupled into complementary fluids from joint and bone. These two streams are processed separately and fused at a later stage to capture spatial dependencies and mitigate \textit{collinearity} issues.
    \item \textbf{Temporal dimension.} MGR is introduced to process joints and bones independently. MGR captures action tokens along the temporal dimension and encodes them into Gaussian distributions, thereby modeling temporal dynamics and alleviating noise from pose estimation.
    \item \textbf{Feature Fusion.} Different fusion strategies used to integrate dual-stream features.
    \item \textbf{Downstream Tasks.} Two downstream tasks are introduced, including report generation and action classification. The use of label smoothing loss enhances the generalization performance.
\end{itemize}

\subsection{Multivariate Gaussian Representation}

\noindent \textbf{Rationale.}\label{story:mge} In the field of skeleton-based action recognition, GCN architectures (\textit{e.g.} ST-GCN \cite{yu2018spatio}, CTR-GC \cite{chen2021channel}) typically rely on local convolution in temporal modeling. However, these methods often lack the ability to capture the temporal dynamics of human actions.

Inspired by the breakthroughs in Gaussian splatting \cite{kerbl20233d} from the field of computer graphics, which has demonstrated remarkable performance in rendering tasks, we re-examined the temporal modeling problem for Gaussian splatting. In graphics rendering, Gaussian splatting can represent a highly dense point cloud in a vast spatial domain using only a small number of Gaussian distributions. This insight informs us that an intricate set of original spatial points can, in fact, be effectively described using significantly fewer key points.

We extend this idea to temporal encoding. In many human actions, the movements of keypoints are inherently continuous and can be represented by a compact set of temporal segments. Critical transition points, such as the start and end of accelerations or directional changes, constitute only a small subset relative to the overall temporal sampling space. This observation aligns with early discussions on spatiotemporal interest points (STIP) \cite{laptev2005space}.
Building on these insights, we combined the innovations of STIP with Gaussian splatting, proposing MGR, enabling efficient and expressive temporal encoding.

\noindent\textbf{Input Construction.}
For each joint \( i \), its temporal trajectory is defined as a spatiotemporal point set \( \mathcal{X}_i = \{ \mathbf{x}_{i,t} \}_{t=1}^T \), where \( \mathbf{x}_{i,t} = (x_{i,t}, y_{i,t}, t) \in \mathbb{R}^3 \) contains 2D coordinates and normalized timestamps. To balance spatial and temporal scales, we introduce a time-axis scaling factor \( \alpha \):  
\begin{equation}
\mathbf{x}_{i,t} \leftarrow (x_{i,t}, y_{i,t}, \alpha \cdot t) \quad (\alpha \in \mathbb{R}^+).
\end{equation}

\noindent\textbf{Gaussian Modeling.}
We conjecture that \( \mathcal{X}_i \) is generated by a mixture of \( K \) Gaussian distributions \cite{reynolds2009gaussian}. The probability density function is:  
\begin{equation}
p(\mathbf{x} | \boldsymbol{\theta}_i) = \sum_{k=1}^K \pi_{i,k} \mathcal{N}(\mathbf{x} | \boldsymbol{\mu}_{i,k}, \boldsymbol{\Sigma}_{i,k}),
\end{equation}
where \( \boldsymbol{\theta}_i = \{ \pi_{i,k}, \boldsymbol{\mu}_{i,k}, \boldsymbol{\Sigma}_{i,k} \}_{k=1}^K \), with mixture weights \( \pi_{i,k} \) satisfying \( \sum_{k=1}^K \pi_{i,k} = 1 \). Parameters are optimized using the Expectation Maximization (EM) algorithm.

\begin{figure}
    \centering
    \includegraphics[width=1\linewidth]{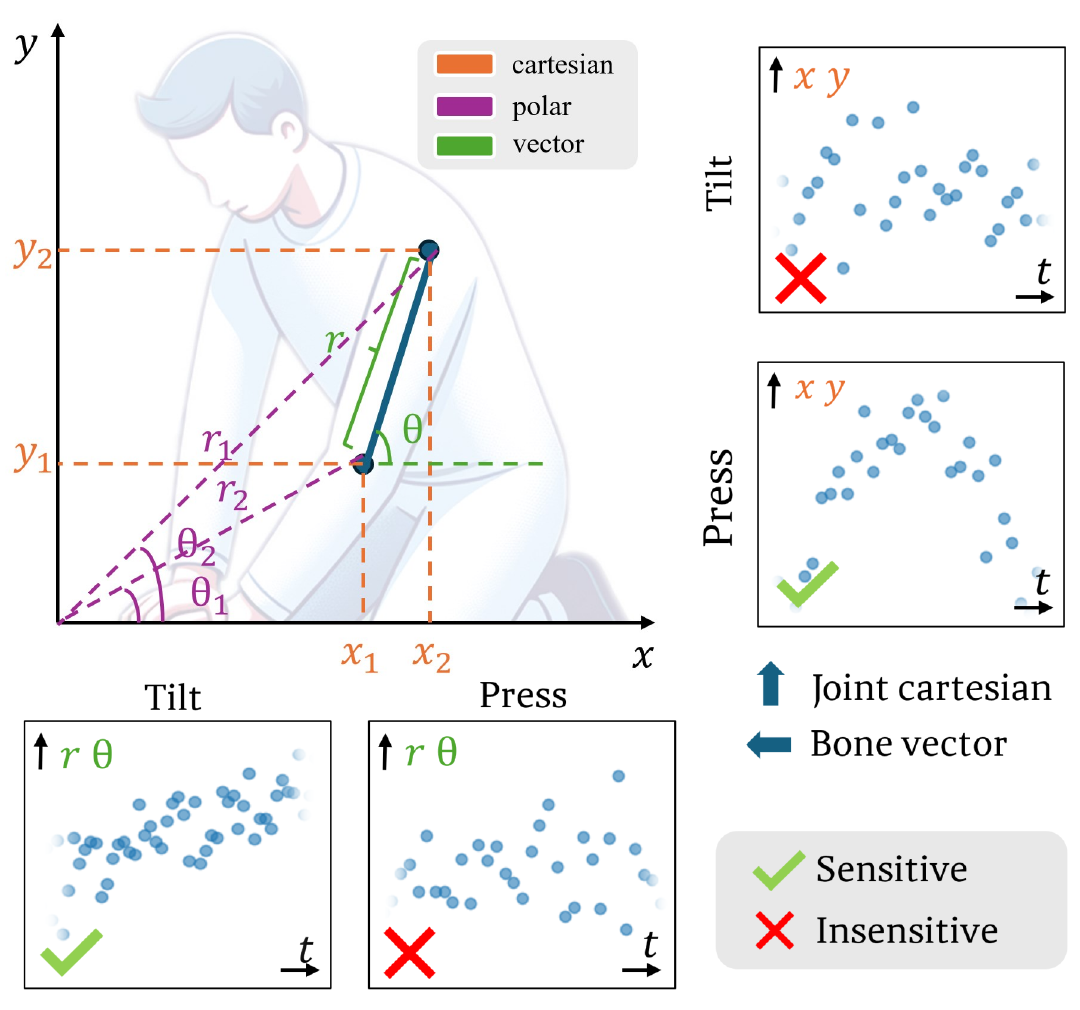}
    \caption{\textbf{Feature Discriminability of Hybrid Spatial Encoding.} The figure illustrates three information modes: cartesian-based, polar-based, and vector-based. Using chest compression and limb tilt as prototypes, the analysis reveals distinct signal-formative capabilities: Sensitive modes generate structured point clusters that fit to kinematic functions, while insensitive modes exhibit noise distributions. The hybrid architecture orchestrates dual-stream processing to adaptively harness these geometric discriminators.}
    \label{fig:sensitive}
\end{figure}

\begin{figure*}[htbp]
    \centering
    \includegraphics[width=\linewidth]{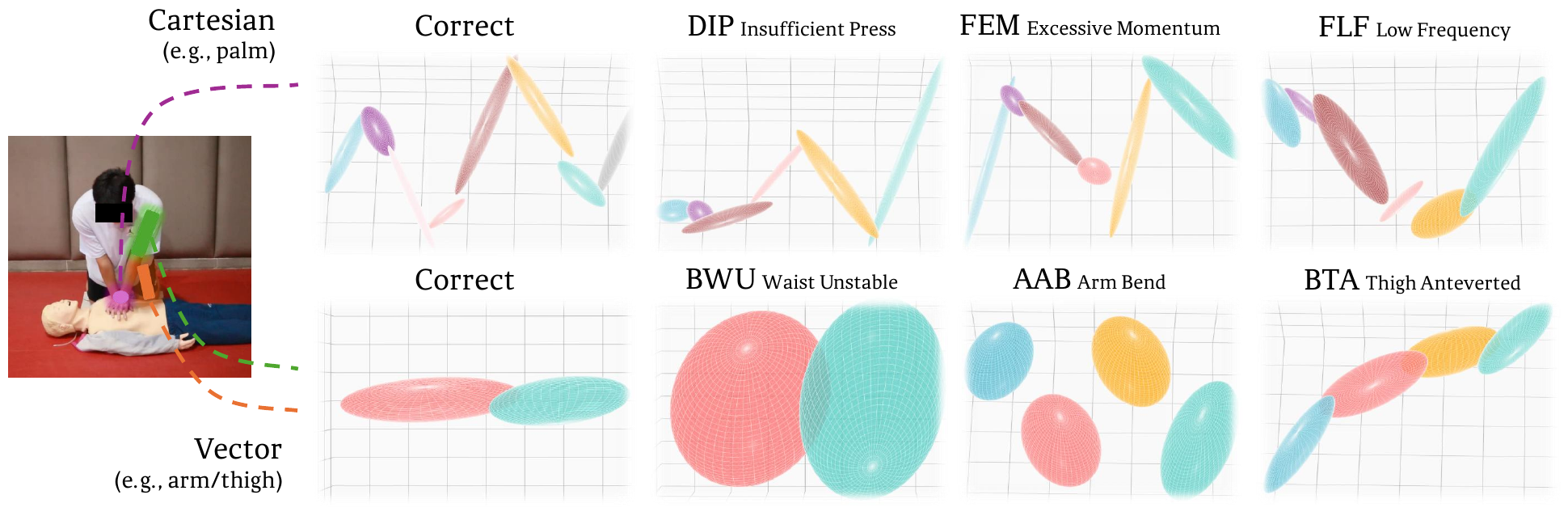}
    \caption{\textbf{Dynamics Visualization of MGR.} Each ellipsoid represents a Gaussian distribution (mean \(\mu\), covariance \(\Sigma\)) of a single joint/bone dynamics over time. From the perspective of stream-specific strengths, Joint Cartesian Stream excels at capturing global trajectory consistency (\textit{e.g.}, palm trajectory); Bone vector better encodes kinematic transitions (\textit{e.g.}, arm bend).}
    \label{fig:dynamics}
\end{figure*}

\noindent\textbf{Action Token.} 
For each Gaussian component \( k \), we extract compressed features from \( \boldsymbol{\mu}_{i,k} \) and \( \boldsymbol{\Sigma}_{i,k} \). Each Gaussian represents an action token. Although we did not use higher dimensions in our architecture (discussed in Section HSE), MGR can be extended to more dimensions, such as incorporating limb angles, states, etc. For the $x$-$y$-$t$ or $r$-$\theta$-$t$ point set, we have a 3D Gaussian distribution: 

\begin{equation}
\mathcal{N}(\mathbf{x}|\mathbf{\mu}, \mathbf{\Sigma}) = \frac{1}{(2\pi)^{d/2} |\mathbf{\Sigma}|^{1/2}} e^{-\frac{1}{2} (\mathbf{x} - \mathbf{\mu})^\top \mathbf{\Sigma}^{-1} (\mathbf{x} - \mathbf{\mu})},
\end{equation}
where, \( \boldsymbol{\mu} \) represents the average position of the joints within this segment, while the covariance matrix  \( \boldsymbol{\Sigma} \) represents the scaling (motion) and rotation (direction) across dimensions.

\noindent\textbf{Covariance Decomposition.} To achieve a representation that is both differentiable and interpretable, we transform the covariance matrix into scale and rotation components that have physical significance \cite{kerbl20233d}.

\begin{equation}
\boldsymbol{\Sigma}_{i,k} = \mathbf{R}_{i,k} \mathbf{S}_{i,k} \mathbf{S}_{i,k}^\top \mathbf{R}_{i,k}^\top,
\end{equation}
where \( \mathbf{S}_{i,k} = \text{diag}(s_{i,k}^x, s_{i,k}^y, s_{i,k}^t) \in \mathbb{R}^{3 \times 3} \) is derived from a 3D vector \( \mathbf{s}_{i,k} \in \mathbb{R}^3 \), \( \mathbf{R}_{i,k} \in \mathbb{R}^{3 \times 3} \) is computed from a unit quaternion \( \mathbf{q}_{i,k} \in \mathbb{R}^4 \).  

This decomposition disentangles the motion dynamics into scale (magnitude of movement) and orientation (direction of movement), aligning with human biomechanical constraints. Finally, we concatenate \( \boldsymbol{\mu}_{i,k} \), \( \mathbf{s}_{i,k} \), and \( \mathbf{q}_{i,k} \) into a compact 10D tensor:  
\begin{equation}
   \mathbf{f}_{i,k} = [\boldsymbol{\mu}_{i,k}; \ \mathbf{s}_{i,k}; \ \mathbf{q}_{i,k}] \in \mathbb{R}^{3 + 3 + 4 = 10}.
\end{equation}

\subsection{Hybrid Spatial Encoding}
\label{sec:spaenc}

\noindent \textbf{Rationale.}
Human action recognition, especially in medical scenarios, is based on modeling anatomical structures from sparse keypoints. Research has shown that 2D point and line combinations provide a strong impression of the type of action \cite{johansson1973visual}.
Both absolute joint positions and relative bone kinematics encode complementary motion semantics \cite{shi2019two}.
The features have two streams of information including direct use of Cartesian coordinates as input (coordinate-based encoding, $(x,y,t)$); and angles or vectors as input (vector-based encoding $(r,\theta,t)$). In terms of anatomy, if bones are projected into a 2D space, information such as angles can be expressed in polar coordinates. These features are critically important in the recognition of medical actions. See Figure \ref{fig:sensitive} for the sensitivity analysis of the two streams.

Let us discuss the limitations of prior encoding schemes:
\begin{itemize}
    \item Joint Cartesian \textit{w/} Bone Cartesian: Widely adopted in GCNs and variants using a dynamic graph or attention mechanism \cite{yu2018spatio,wang2019dynamic,ye2020dynamic,ye2020attention}. This approach forces networks to learn positions, resulting in inefficient feature redundancy.
    \item Joint Polar \textit{w/} Bone Vector: Despite their anatomical grounding, angular wraparound artifacts (\textit{e.g.}, \(-\pi \leftrightarrow \pi\) discontinuities) induce gradient instability, as evidenced by our polar-based model in ablation studies; see supplementary material.
\end{itemize}

We propose the HSE solution that takes advantage of the joint Cartesian coordinates $(x,y,t)$ and the bone vector $(r,\theta,t)$, and decouples absolute localization from relative kinematics, contextualizing fine-grained bone dynamics.

\noindent \textbf{Semantics Disentangling.}
Joint stream (Cartesian encoding) is represented in the \(xyt\)-space to preserve absolute spatial-temporal positions. Bone stream (vector encoding) is parameterized by dynamic vector features in \(r\theta t\)-space.
\begin{equation}
\mathbf{J}_i = [x_i, y_i, t] \in \mathbb{R}^3,
\mathbf{B}_{ij} = [\Delta r_{ij}, \theta_{ij}, t] \in \mathbb{R}^3,
\end{equation}
where \(\Delta r_{ij} = \| \mathbf{J}_j - \mathbf{J}_i \|_2\) and \(\theta_{ij} = \arctan\left(\frac{y_j - y_i}{x_j - x_i}\right)\).
Concatenating raw joint and bone features ($x,y,r,\theta,t$) as input to MGE may risk \textit{multicollinearity}, as Cartesian coordinates and polar parameters are geometrically interdependent (\(x = r\cos\theta, y = r\sin\theta\)).
Our MGE module first processes joints and bones in isolated embedding spaces, disentangling absolute and relative semantics. Subsequent fusion operates on decorrelated high-level features.

\noindent \textbf{Feature Fusion Strategy.}
We adopt fusion with multiple variants: Cross-Attention (Eq. \ref{eq:crossattn}), Interleaved Concatenation (Eq. \ref{eq:interleaved}), and others. Different fusion strategies can work differently on various data complexities and dataset types.

\begin{equation}
\label{eq:crossattn}
    \mathbf{F} = \text{LayerNorm}(\mathbf{F}_J + \text{MultiHead}(\mathbf{Q},\mathbf{K},\mathbf{V})).
\end{equation}
\begin{equation}
\label{eq:interleaved}
\mathbf{F} = \text{Interleave}(\mathbf{F}_J, \mathbf{F}_B) \in \mathbb{R}^{2M \times K \times 10},
\end{equation}
where $\text{Interleave}(\mathbf{F}_J, \mathbf{F}_B)$ rearranges features in an alternating pattern $(j_1, b_1, j_2, b_2, \ldots, j_M, b_M)$ for each skeleton. See the supplementary material for details.

\begin{table*}[htbp]
\centering
\footnotesize
\resizebox{\textwidth}{!}{
\begin{tabular}{@{}lllccccccc@{}}
\toprule
\multirow{2}{*}{\textbf{Model}} & \multirow{2}{*}{Modality} & \multirow{2}{*}{Backbone} & \multirow{2}{*}{Dual-stream} & \multirow{2}{*}{Pre-trained} & \multicolumn{3}{c}{\textbf{Accuracy}} & & \textbf{GFLOPs} \\ \cmidrule(lr){6-8} \cmidrule(l){10-10}
& & & & & Top-1 & Top-5 & Mean & & Per sample \\ \midrule
TSM \cite{lin2019tsm} & RGB & ResNet-50 & \ding{55} & \ding{55} & $0.8665$ & $0.9820$ & $0.8469$ & & $131.83$ \\
TSN \cite{wang2018temporal} & RGB & ResNet-50 & \ding{55} & Kinetics-400 & $0.9009$ & $0.9867$ & $0.8899$ & & $251.12$ \\
TPN \cite{yang2020temporal} & RGB & ResNet-50 & \ding{55} & \ding{55} & $0.8111$ & $0.9727$ & $0.7235$ & & $168.00$ \\
TIN \cite{shao2020temporal} & RGB & ResNet-50 & \ding{55} & \ding{55} & $0.8306$ & $0.9766$ & $0.7951$ & & $131.83$ \\
C3D \cite{tran2015learning} & RGB & 3D ConvNet & \ding{55} &    Sports-1M & \textcolor{orange}{$0.9165$} & \textcolor{blue}{$0.9906$} & $0.9069$ & & $308.92$ \\
I3D \cite{carreira2017quo} & RGB & ResNet-50 & \ding{55} & \ding{55} & $0.8462$ & $0.9789$ & $0.8331$ & & $266.80$ \\
SlowFast \cite{feichtenhofer2019slowfast} & RGB & ResNet-50 & \ding{51} & Kinetics-400 & $0.9144$ & \textcolor{orange}{$0.9874$} & \textcolor{blue}{$0.9107$} & & $97.27$ \\
TimeSformer  \cite{bertasius2021space}& RGB & ViT & \ding{55} & \ding{55} & $0.8283$ & $0.9781$ & $0.8043$ & & $360.89$ \\ \midrule
ST-GCN \cite{yu2018spatio} & Skeleton & GCN & \ding{55} & \ding{55} & $0.8618$ & $0.9758$ & $0.8356$ & & $43.76$ \\
2s-aGCN \cite{shi2019two} & Skeleton & GCN & \ding{51} & \ding{55} & $0.8907$ & $0.9703$ & $0.8812$ & & $50.01$ \\
STGCNPP \cite{duan2022pyskl} & Skeleton & GCN & \ding{55} & \ding{55} & $0.8938$ & $0.9742$ & $0.8777$ & & $21.69$ \\
PoseC3D \cite{duan2022revisiting} & Skeleton & 3D ConvNet & \ding{55} & \ding{55} & $0.8798$ & $0.9727$ & $0.8582$ & & $24.51$ \\ 
InfoGCN \cite{chi2022infogcn}    &  Skeleton & GCN & \ding{55} & \ding{55}  & $0.8977$    & $0.9789$  &  $0.8825$ & & $38.47$ \\
SkateFormer \cite{do2024skateformer} &  Skeleton & Transformer    & \ding{55} & \ding{51}  & $0.9047$    & $0.9789$  &  $0.8907$ & & $42.17$  \\
HDGCN \cite{he2023hdgcn}      &  Skeleton & GCN & \ding{55} & \ding{51}  & $0.9063$    & $0.9797$  &  $0.8911$ & & $39.50$  \\
\midrule
\textbf{Ours} \textit{(MGR w/ HSE)} & Skeleton & CNN & \ding{51} & \ding{55} & \textcolor{blue}{$0.9212$} & $0.9836$ & \textcolor{orange}{$0.9082$} & & \textcolor{orange}{$4.45$} \\
\textbf{Ours} \textit{(MGR only)} & Skeleton & CNN & \ding{55} & \ding{55} & $0.8954$ & $0.9602$ & $0.8836$ & & \textcolor{blue}{{$2.23$}} \\ \bottomrule
\end{tabular}
}
\caption{\textbf{Model Performance Across Modalities.} \textcolor{blue}{Blue} indicates the best result and \textcolor{orange}{Orange} indicates the second-best result.}
\label{tab:baseline}
\end{table*}

\subsection{Temporal Dynamics with MGR}

As illustrated in Figure \ref{fig:dynamics}, our temporal encoding module generates compact yet discriminative representations through MGR. The temporal evolution of actions (\textit{e.g.}, chest compression) is encoded as compact sequences of Gaussian components that represent motion primitives. Remarkably, complex 60 frame motions can only be represented by $\approx6$ Gaussians, demonstrating MGR’s ability to distill high-level motion primitives.
Gaussian parameters (mean \(\mu\), covariance \(\Sigma\)) create separable clusters in feature space, enabling classifiers to achieve 92.1\% accuracy with minimal fine-tuning (see Table \ref{tab:baseline}).

This analysis confirms that Gaussian-based representation learning bridges the gap between raw pose dynamics and specific semantics, an advantage for medical applications that require both precision and interpretability.

\subsection{Downstream Tasks}

\noindent \textbf{Loss Function.}
Taking into account the fine-grained nature of the medical model and the fact that some labels have few or uneven samples, we use MixUp \cite{zhang2017mixup} to comprehensively improve the effectiveness of the model in processing medical data. Through linear interpolation of inputs and labels, the model can be generalized by generating synthetic samples. For a given pair of samples \((x_i, y_i)\) and \((x_j, y_j)\), the mixed input and label are computed as:
\begin{equation}
\tilde{x} = \lambda x_i + (1 - \lambda) x_j, \quad
\tilde{y} = \lambda y_i + (1 - \lambda) y_j,
\end{equation}
where \(\lambda \sim \text{Beta}(\alpha, \alpha)\) for $\alpha\in(0,+\infty)$ is sampled from a Beta distribution, and \(\alpha > 0\) controls the interpolation strength. The MixUp loss is defined as:
\begin{equation}
\mathcal{L}_{\text{MixUp}} = \lambda \mathcal{L}(f(\tilde{x}), y_i) + (1 - \lambda) \mathcal{L}(f(\tilde{x}), y_j).
\end{equation}

By combining pairs of inputs and labels, smoother decision boundaries and reduced overfitting are encouraged, which helps to enhance the robustness of the model. For other loss strategies such as CE and Label Smoothing \cite{muller2019does}.

\noindent\textbf{Classifier.}
The fused features \( \mathbf{F}_{\text{fused}} \) are mapped to category scores through multilayer CNN with spatiotemporal pooling and network outputs \( \mathbf{y}_{\text{pred}} \). See the supplementary material for the detailed architecture.

\noindent \textbf{Evaluation Report Generation.} 
For generating a report of the evaluation, MGR-encoded skeletal tensors undergo spatiotemporal tokenization to bridge visual and textual modalities. Specifically, we first discretize continuous kinematic features (depth, frequency, posture angles) into clinically-grounded linguistic descriptors (``insufficient 4cm compression",``optimal 110bpm rhythm") using quantization bins derived from AHA guidelines, then made final inferences (metrics$\rightarrow$manifestation$\rightarrow$consequence) through a logical chain reasoning mechanism based on dataset analysis. Effectiveness score calculation and causal analysis is
shown in the supplementary material.

\section{Experiments}

All experiments were implemented in PyTorch and use the MMPose framework \cite{mmpose2020}. To generate a pose input for \textsc{GaussMedAct} from RGB modality, we adopted RTMpose \cite{jiang2023rtmpose}. The image sequences are uniformly sampled to 32-frame clips with a spatial resolution of 224×224. To ensure fairness, all models shared identical training-test splits (80\%-20\% random partition) and underwent rigorous adjustments for confounding factors, as detailed in Table \ref{tab:baseline}. All skeleton baselines were trained using the exact same skeleton data. We trained models using early stopping with a maximum of 300 epochs. Memory-intensive models were deployed on NVIDIA A100 GPUs, while other models utilized NVIDIA A6000 GPUs.

\begin{table}[htbp]
\footnotesize
\centering
\begin{tabular}{ccccc}
\toprule
\multirow{2}{*}{\textbf{Model}} & \multicolumn{1}{l}{\multirow{2}{*}{Modality}}& \multirow{2}{*}{Epoch} & \multicolumn{2}{c}{\textbf{Accuracy}} \\ \cmidrule(lr){4-5} 
                       & \multicolumn{1}{l}{}                         &                        & Top-1      & Top-3      \\ \midrule
TSN-pretrained & RGB      & 50          & $0.9067$ & $0.9921$ \\
TSN            & Flow     & 50          & $0.8304$ & $0.9851$ \\ \cmidrule{2-5} 
STGCN-best     & Skeleton & 50          & \textcolor{orange}{$0.9246$} & \textcolor{blue}{$0.9970$} \\
PoseC3D        & Skeleton & 240          & $0.9208$ & $0.9922$ \\ \midrule
\textbf{Ours}  & Skeleton & 100         & \textcolor{blue}{$0.9524$} & \textcolor{orange}{$0.9950$}  \\ \bottomrule
\end{tabular}
\caption{\textbf{Cross-dataset Evaluation on \textit{Coach}.} The dataset is a medical action dataset with 14 categories. \textcolor{blue}{Blue} indicates the best result and \textcolor{orange}{Orange} indicates the second-best.}
\label{tab:coach}
\end{table}

\subsection{Comparative Analysis}
We conduct comprehensive evaluations of multimodal approaches on the \textsc{CPREval-6K} dataset, comparing our skeleton-based \textsc{GaussMedAct} with state-of-the-art methods that use RGB and skeleton modalities. Four metrics are adopted: Top-1/5 accuracy, class-wise mean accuracy, and computational complexity (GFLOPs). 
The top-1 accuracy receives the primary emphasis due to its clinical relevance as it critically affects CPR effectiveness. 
There may be multiple relatively reasonable answers (\textit{e.g.}, when both primary and secondary labels exist, and the model identifies the secondary category). In such cases, using the Top-5 accuracy metric would provide a fairer evaluation of the model's performance.
Class-wise mean accuracy ensures balanced performance across CPR errors, particularly for low-frequency but high-risk categories (\textit{e.g.}, depth-excess).
As shown in Table \ref{tab:baseline} and Figure \ref{fig:baseline}, three key observations emerge:

\noindent\textbf{Modality.} Skeleton-based models demonstrate significantly lower computational requirements ($6.13\times$ fewer GFLOPs on average) compared to RGB counterparts. Even when accounting for pose extraction costs (4.03-5.45 GFLOPs for RTMpose), the total computation remains substantially lower than RGB methods. However, RGB approaches achieve higher accuracy ceilings (91.65\% vs. 89.38\% for prior skeleton methods), indicating that the RGB modality has specific advantages in feature richness.

\noindent \textbf{Superiority of \textsc{GaussMedAct}.} Our method establishes new state-of-the-art performance with 92.12\% Top-1 accuracy, surpassing all existing models, including RGB-based approaches. In particular, this is achieved with only 4.45 GFLOPs, which is 10\% of the computational cost required by ST-GCN (43.76 GFLOPs). The accuracy improvement over RGB models confirms the underexploited potential of skeleton data in medical action recognition when coupled with effective representation learning.

\noindent\textbf{Efficiency-Performance Trade-off.} The \textit{MGR-only} variant (89.54\% Top-1) already outperforms all skeleton baselines while requiring only half the computational complexity of full model, validating the efficacy of MGR. The complete model further improves performance by 2.58\%, demonstrating synergistic effects between MGR and HSE.

\subsection{Robustness}
\noindent\textbf{Cross-Dataset Evaluation.} 
We conduct rigorous cross-dataset validation using the recently released medical CPR-Coach benchmark \cite{Wang_2024_CVPR}, which contains 14 classes. Following the official (60/40) split protocol, we compare \textsc{GaussMedAct} with four approaches in different modality. Training configurations strictly follow original papers for all methods. All models use the best results reported under different settings in the original paper.
As shown in Table \ref{tab:coach}, our method achieves 2.78\% absolute improvement in Top-1 accuracy while maintaining real-time performance. 

\noindent\textbf{Perturbation and real scene.}
To evaluate the model's robustness, particularly against occlusion and sensor variations, we introduced an additional test set comprising 114 real-world training videos. These videos, captured primarily from beginners, feature realistic scenarios including occlusions and diverse mobile phone sensors. Other experiments are provided in the supplementary materials.

\begin{figure}
    \centering
    \includegraphics[width=1\linewidth]{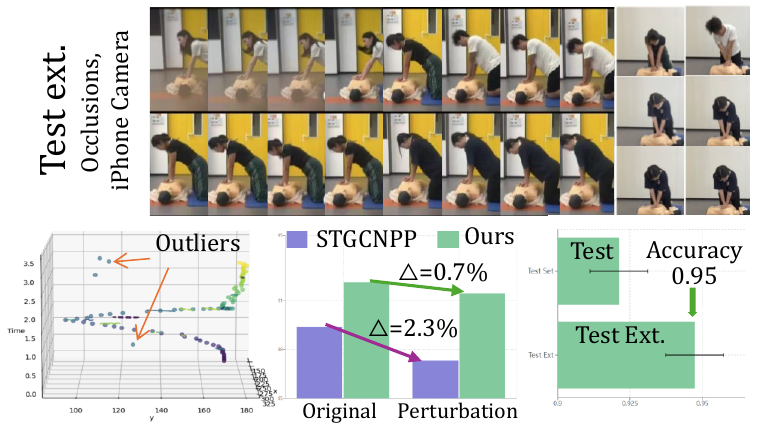}
    \caption{\textbf{Test Extention for Real Scene.} \textsc{GaussMedAct} shows inherent robustness to perturbations comparing to STGCNPP. In practical usage scenarios, the accuracy is higher than \textsc{CPREval-6k}.}
    \label{fig:placeholder}
\end{figure}

\section{Social Impact}

Beyond technical, our framework has demonstrated tangible social impact through deployment in CPR training programs. Partnering with training centers, \textsc{GaussMedAct} has been integrated into standardized courses across several institutions. Quantitative evaluations show a 32\% improvement in practical assessment scores among trainees compared to traditional methods under identical conditions, while instructors highlights enhanced diagnosis precision and actionable feedback. These real-world validations underscore the potential of AI-assisted medical training to improve resuscitation quality and patient outcomes.

\section{Conclusions}
\label{sec:conc}

This paper addresses critical dataset and methodology gaps in medical action evaluation. We introduce \textsc{CPREval-6k}, a much-needed fine-grained medical dataset that incorporates expert-validated hierarchical annotations which capture subtle error patterns.
Through comprehensive comparative studies, we demonstrate that our proposed framework using spatiotemporal Gaussian mixture representation in decoupled joint and bone spaces outperforms both RGB- and skeleton-based models in accuracy while achieving significant computational cost reduction. The framework also exhibits robustness to cross-dataset generalization.
These contributions establish a new foundation for the real-time CPR evaluation, with applications that extend to other medical assessments.

\bibliography{aaai2026}

\end{document}